\documentclass[10pt,twocolumn,letterpaper]{article}

\usepackage[pagenumbers]{cvpr} 

\usepackage[dvipsnames]{xcolor}

\definecolor{cvprblue}{rgb}{0.21,0.49,0.74}
\usepackage[pagebackref,breaklinks,colorlinks,citecolor=cvprblue]{hyperref}

\def\paperID{11433} 
\def\confName{CVPR}
\def\confYear{2024}

\usepackage{graphicx}
\usepackage{amsmath}
\usepackage{amssymb}
\usepackage{booktabs}

\usepackage{bm}
\usepackage{url}
\usepackage{pifont}
\usepackage{bbding}
\usepackage{comment}
\usepackage{caption}
\usepackage{multirow}
\usepackage{makecell}
\usepackage{amsfonts}
\usepackage{float}
\usepackage{multicol}
\usepackage{ragged2e}
\usepackage[misc]{ifsym}
\usepackage{pifont}
\usepackage{dsfont}
\usepackage[hang,flushmargin]{footmisc}

\usepackage{xcolor,colortbl}
\newcommand{\cmark}{\ding{51}}%
\newcommand{\xmark}{\ding{55}}%
\usepackage[accsupp]{axessibility} 

\def\ie{{\em i.e. }}
\def\eg{{\em e.g. }}
\def\etal{{\em et al. }}

\DeclareRobustCommand{\name}{InstaGen}

\title{InstaGen: Enhancing Object Detection by Training on Synthetic Dataset\vspace{-8pt}}

\author{
Chengjian Feng$^{1}$
\quad
Yujie Zhong$^{1}$
\quad
Zequn Jie$^{1,\dagger}$
\quad
Weidi Xie$^{2,\dagger}$
\quad
Lin Ma$^{1}$\\[1pt]
$^{1}$ Meituan Inc.
\quad
$^{2}$ CMIC, Shanghai Jiao Tong University 
\\[1pt]
fcjian@outlook.com \quad jaszhong@hotmail.com \quad zequn.nus@gmail.com
\\[1pt]
weidi@sjtu.edu.cn \quad forest.linma@gmail.com
\\[1pt]
\url{https://fcjian.github.io/InstaGen}
\vspace{-4pt}
}
\newcommand\blfootnote[1]{%
    \begingroup 
    \renewcommand\thefootnote{}\footnote{#1}%
    \addtocounter{footnote}{-1}%
    \endgroup 
}

\begin{document}

\twocolumn[{%
\renewcommand\twocolumn[1][]{#1}%
\maketitle
\maketitle

\vspace{-0.9cm}
\begin{center}
   \centering
   \includegraphics[width=1\linewidth]{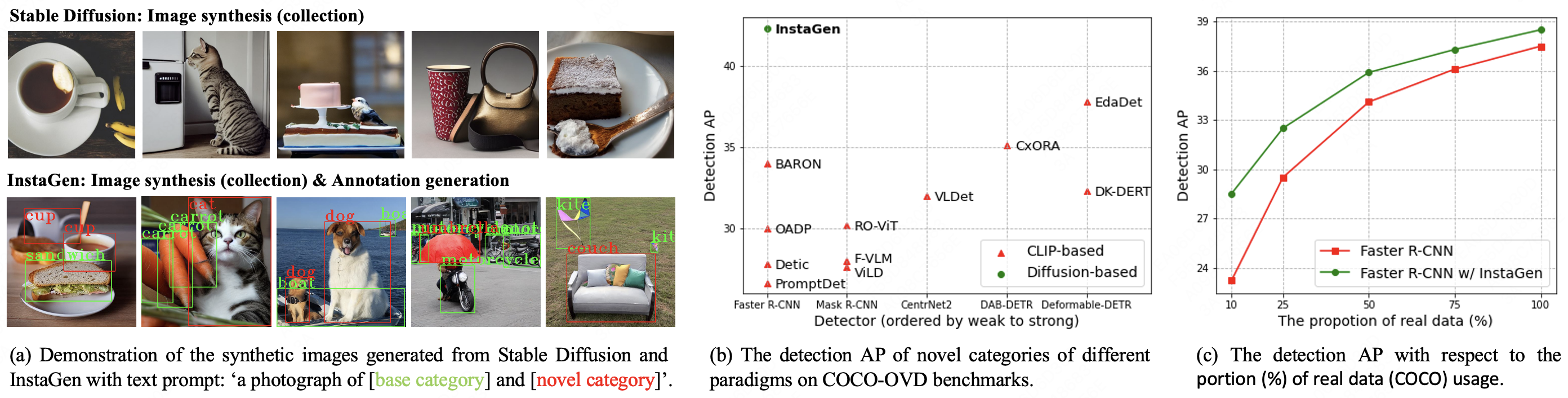}
   \vspace{-16pt}
   \captionof{figure}{
   (a) The synthetic images generated from \textbf{Stable Diffusion} and our proposed \textbf{\name}, which can serve as a {\em dataset synthesizer} for sourcing photo-realistic images and instance bounding boxes at scale. (b) On open-vocabulary detection, 
   training on synthetic images demonstrates significant improvement over CLIP-based methods on novel categories. (c) Training on the synthetic images generated from \name~also enhances the detection performance in close-set scenario, particularly in data-sparse circumstances.
   }
   \vspace{-0.1cm}
   \label{figure:first-page}
\end{center}%
}]

\noindent \blfootnote{$\dagger$: corresponding author.}

\vspace{-8pt}
\begin{abstract}
\vspace{-5pt}
In this paper, we present a novel paradigm to enhance the ability of object detector, {\em e.g.},~expanding categories or improving detection performance, by training on \textbf{synthetic dataset} generated from diffusion models. 
Specifically, we integrate an instance-level grounding head into a pre-trained, generative diffusion model, to augment it with the ability of localising instances in the generated images. The grounding head is trained to align the text embedding of category names with the regional visual feature of the diffusion model, using supervision from an off-the-shelf object detector, and a novel self-training scheme on (novel) categories not covered by the detector. 
We conduct thorough experiments to show that, 
this enhanced version of diffusion model, termed as \textbf{\emph{\name}}, can serve as a data synthesizer,
to enhance object detectors by training on its generated samples, demonstrating superior performance over existing state-of-the-art methods in open-vocabulary ($+4.5$ AP) and data-sparse ($+1.2\sim5.2$ AP) scenarios.
\vspace{-10pt}
\end{abstract}
    
\section{Introduction}
\label{sec:intro}
Object detection has been extensively studied in the field of computer vision, focusing on the localization and categorization of objects within images~\cite{redmon2016you,ren2015faster,he2017mask,feng2021tood,feng2022promptdet}. The common practise is to train the detectors on large-scale image datasets, such as MS-COCO~\cite{lin2014microsoft} and Object365~\cite{shao2019objects365}, where objects are exhaustively annotated with bounding boxes and corresponding category labels. However, the procedure for collecting images and annotations is often laborious and time-consuming, limiting the datasets' scalability.

In the recent literature, text-to-image diffusion models have demonstrated remarkable success in generating high-quality images~\cite{saharia2022photorealistic,rombach2022high}, 
that unlocks the possibility of training vision systems with synthetic images. In general, existing text-to-image diffusion models are capable of synthesizing images based on some free-form text prompt, as shown in the first row of Figure~\ref{figure:first-page}a. 
Despite being photo-realistic, such synthesized images {\em can not} support training sophisticated systems, that normally requires the inclusion of instance-level annotations, 
{\em e.g.}, bounding boxes for object detection in our case. 
In this paper, we investigate a novel paradigm of 
{\em dataset synthesis} for training object detector, 
{\em i.e.}, augmenting the text-to-image diffusion model to generate instance-level bounding boxes along with images.

To begin with, we build an image synthesizer by fine-tuning the diffusion model on existing detection dataset. 
This is driven by the observation that off-the-shelf diffusion models often generate images with only one or two objects on simplistic background, training detectors on such images may thus lead to reduced robustness in complex real-world scenarios. 
Specifically, we exploit the existing detection dataset, and subsequently fine-tune the diffusion model with the image-caption pairs, constructed by taking random image crops, and composing the category name of the objects in the crop. As illustrated in the second row of the Figure~\ref{figure:first-page}a, once finetuned, the image synthesizer now enables to produce images with multiple objects and intricate contexts, thereby providing a more accurate simulation of real-world detection scenarios.

To generate bounding boxes for objects within synthetic images, 
we propose an instance grounding module that establishes the correlation between the regional visual features from diffusion model and the text embedding of category names, and infers the coordinates for the objects' bounding boxes.
Specifically, we adopt a two-step training strategies,
{\em firstly}, we train the grounding module on synthetic images, with the supervision from an off-the-shelf object detector, which has been trained on a set of base categories; 
{\em secondly}, we utilize the trained grounding head to generate pseudo labels for a larger set of categories, including those not seen in existing detection dataset, and self-train the grounding module. Once finished training, the grounding module will be able to identify the objects of arbitrary category and their bounding boxes in the synthetic image, by simply providing the name in free-form language. 

To summarize, we explore a novel approach to enhance object detection capabilities, such as expanding detectable categories and improving overall detection performance, by training on {\em synthetic dataset} generated from diffusion model.
We make the following contribution:
(i) We develop an image synthesizer by fine-tuning the diffusion model, with image-caption pairs derived from existing object detection datasets, our synthesizer can generate images with multiple objects and complex contexts, offering a more realistic simulation for real-world detection scenarios.
(ii) We introduce a data synthesis framework for detection, termed as \textbf{{\name}}. This is achieved through a novel grounding module that enables to generate labels and bounding boxes for objects in synthetic images.
(iii) We train standard object detectors on the combination of {\em real and synthetic} dataset, and demonstrate superior performance over existing state-of-the-art detectors across various benchmarks, including open-vocabulary detection (increasing Average Precision [AP] by +4.5), data-sparse detection (enhancing AP by +1.2 to +5.2), and cross-dataset transfer (boosting AP by +0.5 to +1.1).

\section{Related Work}
\label{sec:related}

\noindent \textbf{Object Detection.}
Object detection aims to simultaneously predict the category and corresponding bounding box for the objects in the images. Generally, object detectors~\cite{redmon2016you,ren2015faster,feng2021tood,feng2021exploring,feng2023aedet} are trained on a substantial amount of training data with bounding box annotations and can only recognize a predetermined set of categories present in the training data. In the recent literature, to further expand the ability of object detector, open-vocabulary object detection (OVD) has been widely researched, for example, OVR-CNN~\cite{zareian2021open} introduces the concept of OVD and pre-trains a vision-language model with image-caption pairs. The  subsequent works make use of the robust multi-modal representation of CLIP~\cite{radford2021learning}, 
and transfer its knowledge to object detectors through knowledge distillation~\cite{gu2021open,xie2021zsd}, exploiting extra data~\cite{feng2022promptdet,zhou2022detecting} 
and text prompt tuning~\cite{du2022learning,feng2022promptdet}. 
In this paper, we propose to expand the ability of object detectors, {\em e.g.}, expanding categories
or improving detection performance, 
by training on synthetic dataset. 

\vspace{3pt}\noindent \textbf{Generative Models.}
Image generation has been considered as a task of interest in computer vision for decades.
In the recent literature, significant progress has been made, 
for example, the generative adversarial networks (GANs)~\cite{goodfellow2020generative}, 
variational autoencoders (VAEs)~\cite{kingma2013auto}, 
flow-based models~\cite{kingma2018glow}, 
and autoregressive models (ARMs)~\cite{van2016conditional}. 
More recently, there has been a growing research interest in diffusion probabilistic models (DPMs), which have shown great promise in generating high-quality images across diverse datasets. For examples, GLIDE~\cite{nichol2021glide} utilizes a pre-trained language model and a cascaded diffusion structure for text-to-image generation. DALL-E~2~\cite{ramesh2022hierarchical} is trained to generate images by inverting the CLIP image space, while Imagen~\cite{saharia2022photorealistic} explores the advantages of using pre-trained language models. Stable Diffusion~\cite{rombach2022high} proposes the diffusion process in VAE latent spaces rather than pixel spaces, effectively reducing resource consumption. In general, the rapid development of generative models opens the possibility for training large models with synthetic dataset. 

\begin{figure*}[ht]
  \centering
  \begin{subfigure}[b]{0.48\textwidth}
    \centering
    \includegraphics[height=4.4cm]{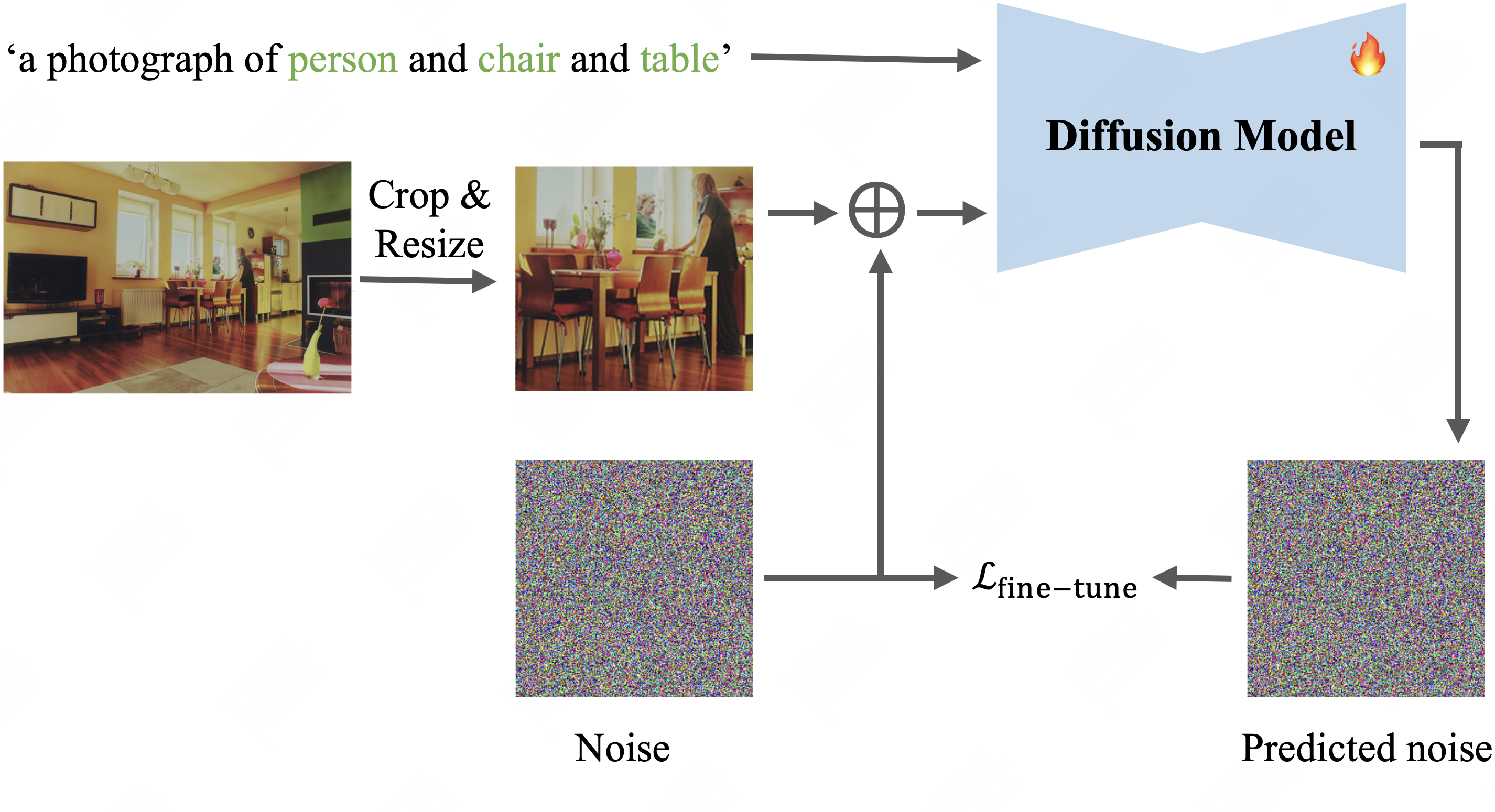}
    \caption{Fine-tuning diffusion model on detection dataset.}
    \label{figure:finetune-a}
  \end{subfigure}
  \hfill
  \begin{subfigure}[b]{0.48\textwidth}
    \centering
    \includegraphics[height=4.8cm]{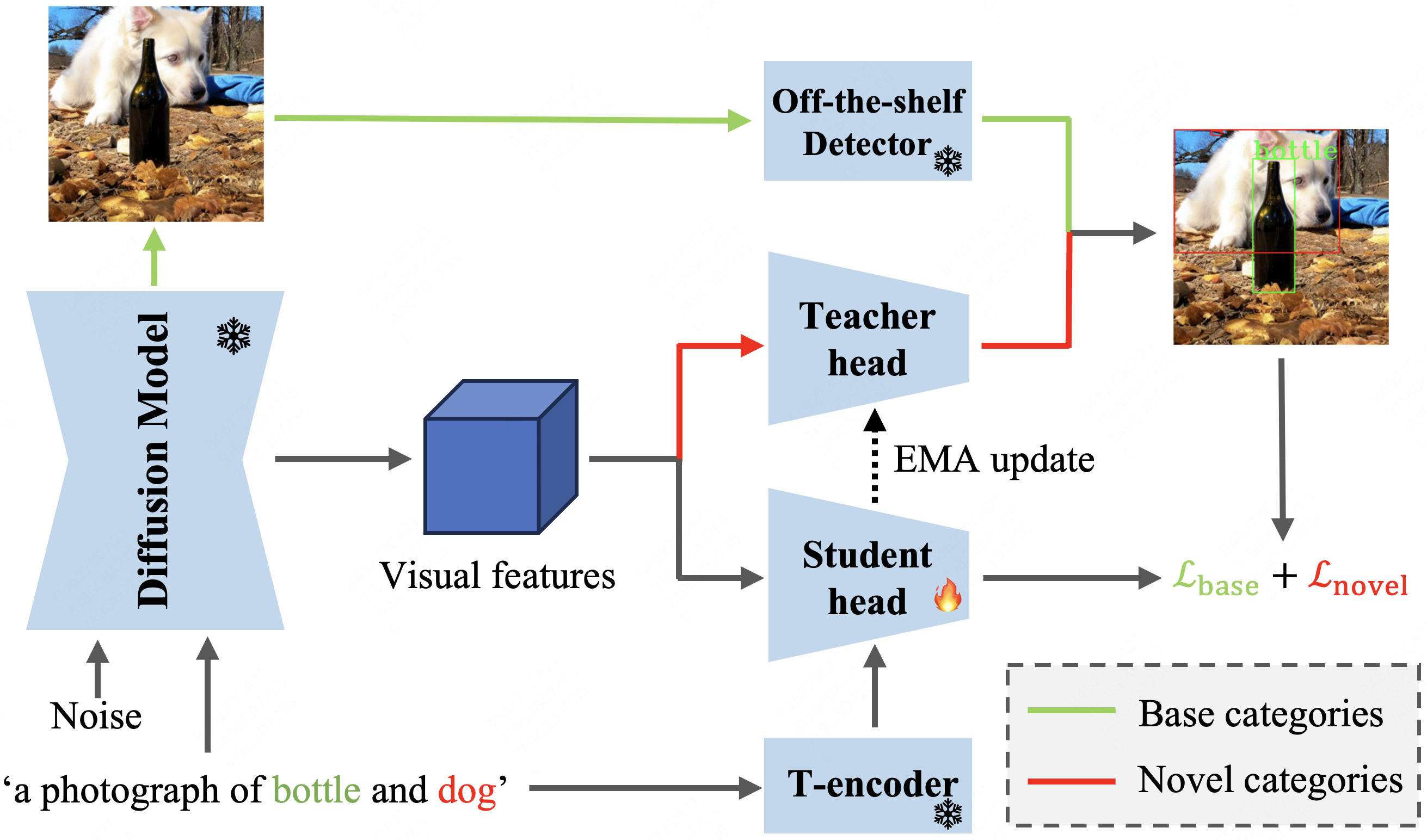}
    \caption{Supervised training and self-training for grounding head (\ie student).}
    \label{figure:finetune-b}
  \end{subfigure}
  \vspace{-5pt}
  \caption{Illustration of the process for finetuning diffusion model and training the grounding head: (a) stable diffusion model is fine-tuned on the detection dataset on base categories. (b) The grounding head is trained on synthetic images, with supervised learning on base categories and self-training on novel categories.}
  \vspace{-14pt}
  \label{figure:finetune}
\end{figure*}

\section{Methodology}
\label{sec:methodology}

In this section, we present details for constructing a {\em dataset synthesizer}, that enables to generate photo-realistic images with bounding boxes for each object instance, and train an object detector on the combined real and synthetic datasets.

\vspace{-1mm}
\subsection{Problem Formulation}
Given a detection dataset of real images with manual annotations,
{\em i.e.}, $\mathcal{D}_{\text{real}} = \{(x_1, \mathcal{B}_1, \mathcal{Y}_1), \dots, (x_N, \mathcal{B}_N, \mathcal{Y}_N)\}$, where $\mathcal{B}_i = \{b_1, \dots, b_m | b_j \in \mathbb{R}^{2 \times 2}\}$ denotes the set of box coordinates for the annotated instances in one image, and $\mathcal{Y}_i = \{y_1, \dots, y_m | y_j \in \mathcal{R}^{\mathcal{C}_{\text{base}}}\}$ refers to the categories of the instances. Our goal is thus to exploit the given real dataset~($\mathcal{D}_{\text{real}}$), 
to steer a generative diffusion model into {\em dataset synthesizer}, that enables to augment the existing detection dataset, {\em i.e.}, $\mathcal{D}_{\text{final}} = \mathcal{D}_{\text{real}} + \mathcal{D}_{\text{syn}}$. As a result, detectors trained on the combined dataset demonstrate enhanced ability, {\em i.e.}, extending the detection categories or improving the detection performance. 

In the following sections, 
we first describe the procedure for constructing an {\em image synthesizer}, that can generate images suitable for training object detector~(Section \ref{subsec:tune}). To simultaneously generate the images and object bounding boxes, we propose a novel instance-level grounding module, which aligns the text embedding of category name with the regional visual features from {\em image synthesizer}, and infers the coordinates for the objects in synthetic images. To further improve the alignment towards objects of arbitrary category, we adopt self-training to tune the grounding module on object categories not existing in $\mathcal{D}_{\text{real}}$~(Section~\ref{subsec:dataset}). 
As a result, the proposed model, termed as \textbf{\name}, can automatically generate images along with bounding boxes for object instances, and construct {\em synthetic dataset}~($\mathcal{D}_{\text{syn}}$) at scale, leading to improved ability when training detectors on it~(Section~\ref{subsec:detector}).

\subsection{Image Synthesizer for Object Detection}
\label{subsec:tune}
Here, we build our {\em image synthesizer} based on an off-the-shelf stable diffusion model~(SDM~\cite{rombach2022high}). Despite of its impressive ability in generating photo-realistic images, it often outputs images with only one or two objects on simplistic background with the text prompts, for example, `a photograph of a [category1 name] and a [category2 name]’, as demonstrated in Figure~\ref{figure:image-bbox-b}. As a result, object detectors trained on such images may exhibit reduced robustness when dealing with complex real-world scenarios. To bridge such domain gap, we propose to construct the {\em image synthesizer} by fine-tuning the SDM with an existing real-world detection dataset~($\mathcal{D}_{\text{real}}$). 

\vspace{3pt}
\noindent \textbf{Fine-tuning procedure.} 
To fine-tune the stable diffusion model~(SDM), 
one approach is to na\"ively use the sample from detection dataset, for example, randomly pick an image and construct the text prompt with all categories in the image. 
However, as the image often contains multiple objects, 
such approach renders significant difficulty for fine-tuning the SDM, especially for small or occluded objects. 
We adopt a mild strategy by taking random crops from the images, and construct the text prompt with categories in the image crops, as shown in Figure~\ref{figure:finetune-a}.
If an image crop contains multiple objects of the same category, 
we only use this category name once in the text prompt.

\begin{figure*}[t]
    \centering
    \includegraphics[width=\textwidth]{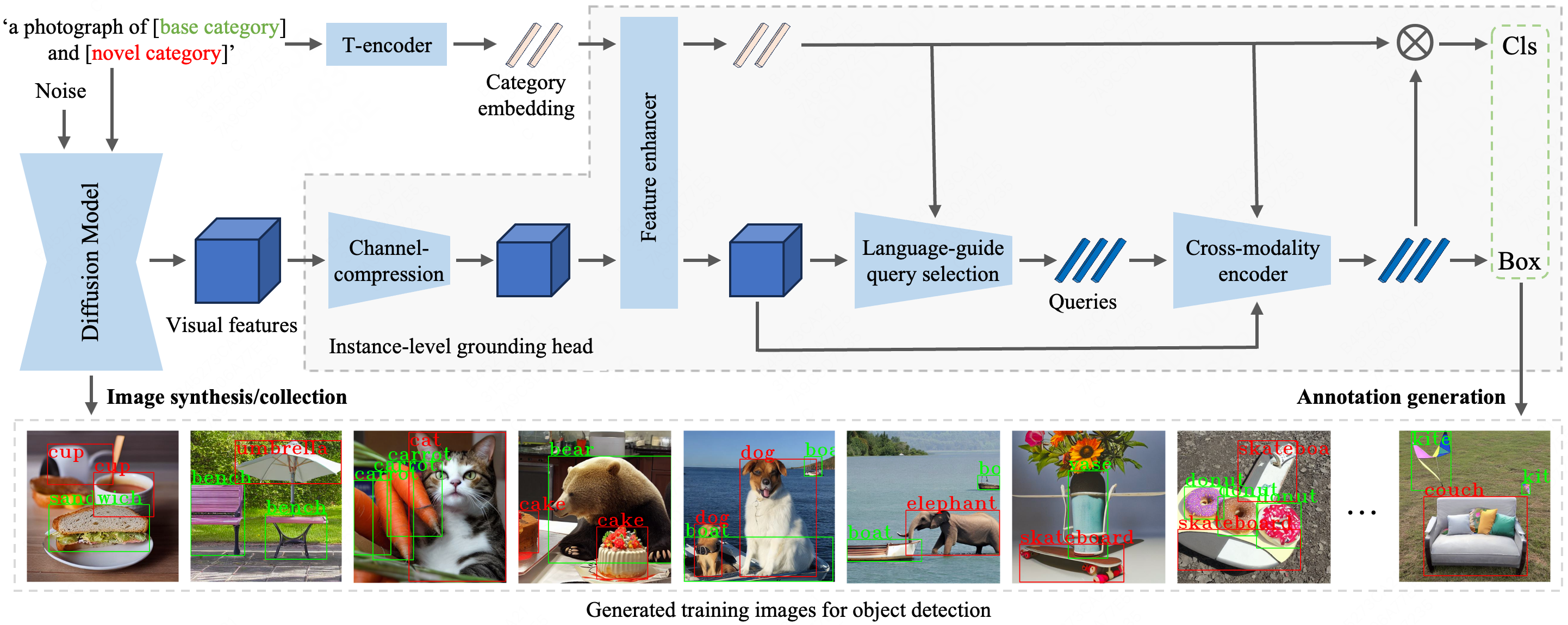}
    \vspace{-.6cm}
    \caption{Illustration of the dataset generation process in \name. The data generation process consists of two steps: (i) \textbf{Image collection}: given a text prompt, SDM generates images with the objects described in the text prompt; (ii) \textbf{Annotation generation}: the instance-level grounding head aligns the category embedding with the visual feature region of SDM, generating the corresponding object bounding-boxes.}
    \vspace{-3mm}
    \label{figure:structure}
\end{figure*}

\vspace{3pt}\noindent \textbf{Fine-tuning loss.} 
We use the sampled image crop and constructed text prompt to fine-tune SDM with a squared error loss on the predicted noise term as follows:
\vspace{-8pt}
\begin{equation}
\mathcal{L}_{\text{fine-tune}} = \mathbb{E}_{z, \epsilon \sim \mathcal{N}(0,1),t,y} \Big[||\epsilon - \epsilon_{\theta}(z^{t},t,y)||_{2}^{2} \Big],
\end{equation}
where $z$ denotes a latent vector mapped from the input image with VAE, $t$ denotes the denoising step, uniformly sampled from $\{1,\dots,T\}$, $T$ refers to the length of the diffusion Markov chain, and $\epsilon_{\theta}$ refers to the estimated noise from SDM with parameters $\theta$ being updated. 
We have experimentally verified the necessity of this fine-tuning step, as shown in Table~\ref{table:componets}.

\subsection{Dataset Synthesizer for Object Detection}
\label{subsec:dataset}
In this section, we present details for steering the {\em image synthesizer} into {\em dataset synthesizer} for object detection, which enables to simultaneously generate images and object bounding boxes. Specifically, we propose an instance-level grounding module that aligns the text embedding of object category, with the regional visual feature of the diffusion model, and infers the coordinates for bounding boxes, effectively augmenting the {\em image synthesizer} with instance grounding, as shown in Figure~\ref{figure:structure}. 
To further improve the alignment in large visual diversity, 
we propose a self-training scheme that enables the grounding module to generalise towards arbitrary categories, including those not exist in real detection dataset~($\mathcal{D}_{\text{real}}$). As a result, our {\em data synthesizer}, termed as \textbf{\name}, can be used to construct synthetic dataset for training object detectors.

\vspace{-8pt}
\subsubsection{Instance Grounding on Base Categories}
\label{subsec:head}
\vspace{-5pt}

To localise the object instances in synthetic images, we introduce an open-vocabulary grounding module, that aims to simultaneously generate image~($x$) and the corresponding instance-level bounding boxes~($\mathcal{B}$) based on a set of categories~($\mathcal{Y}$), {\em i.e.}, $\{x, \mathcal{B}, \mathcal{Y}\} = \Phi_{\text{\name}}(\epsilon, \mathcal{Y})$, where $\epsilon \sim \mathcal{N}(0,I)$ denotes the sampled noise.

To this end, we propose an instance grounding head, 
as shown in Figure~\ref{figure:structure},
it takes the intermediate representation from {\em image synthesizer} and the text embedding of category as inputs, 
then predicts the corresponding object bounding boxes, 
{\em i.e.}, $\{\mathcal{B}_i, \mathcal{Y}_i\} = \Phi_{\text{g-head}}(\mathcal{F}_{i},  \Phi_{\text{t-enc}}(g(\mathcal{Y}_i)))$,
where $\mathcal{F}_i = \{f_i^1, \dots, f_i^n\}$ refers to the multi-scale dense features from the {\em image synthesizer} at time step $t=1$, $g(\cdot)$ denotes a template that decorates each of the visual categories in the text prompt, {\em e.g.}, `a photograph of [category1 name] and [category2 name]', $\Phi_{\text{t-enc}}(\cdot)$ denotes the text encoder.

Inspired by GroundingDINO~\cite{liu2023grounding}, 
our grounding head $\Phi_{\text{g-head}}(\cdot)$ mainly contains four components: (i) a channel-compression layer, implemented with a 3×3 convolution, for reducing the dimensionality of the visual features;
(ii) a feature enhancer, consisting of six feature enhancer layers, to fuse the visual and text features. Each layer employs a deformable self-attention to enhance image features, a vanilla self-attention for text feature enhancers, an image-to-text cross-attention and a text-to-image cross-attention for feature fusion;
(iii) a language-guided query selection module for query initialization. This module predicts top-$N$ anchor boxes based on the similarity between text features and image features. Following DINO~\cite{zhang2022dino}, it adopts a mixed query selection where the positional queries are initialized with the anchor boxes and the content queries remain learnable;
(iv) a cross-modality decoder for classification and box refinement. It comprises six decoder layers, with each layer utilizing a self-attention mechanism for query interaction, an image cross-attention layer for combining image features, and a text cross-attention layer for combining text features.
Finally, we apply the dot product between each query and the text features, followed by a Sigmoid function to predict the classification score $\hat{s}$ for each category. Additionally, the object queries are passed through a Multi-Layer Perceptron (MLP) to predict the object bounding boxes $\hat{b}$, as shown in Figure~\ref{figure:structure}.
We train the grounding head by aligning the category embedding with the regional visual features from diffusion model, as detailed below. \emph{Once trained, the grounding head is open-vocabulary}, 
{\em i.e.}, given any categories (even beyond the training categories), the grounding head can generate the corresponding bounding-boxes for the object instances.

\begin{figure*}[ht]
  \centering
  \begin{subfigure}[b]{0.315\textwidth}
    \centering
    \includegraphics[width=\textwidth]{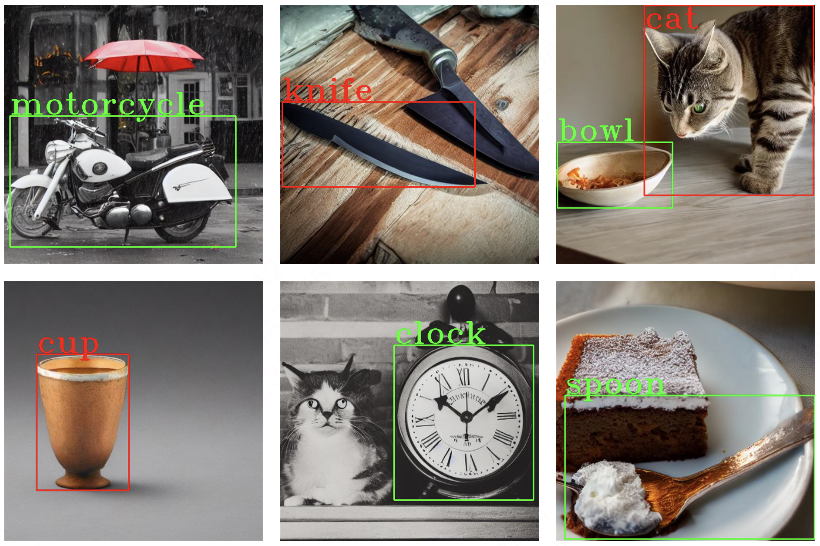}
    \caption{Stable Diffusion + Grounding head w/ \textbf{Supervised training}.}
    \label{figure:image-bbox-a}
  \end{subfigure}
  \hfill
  \begin{subfigure}[b]{0.315\textwidth}
    \centering
    \includegraphics[width=\textwidth]{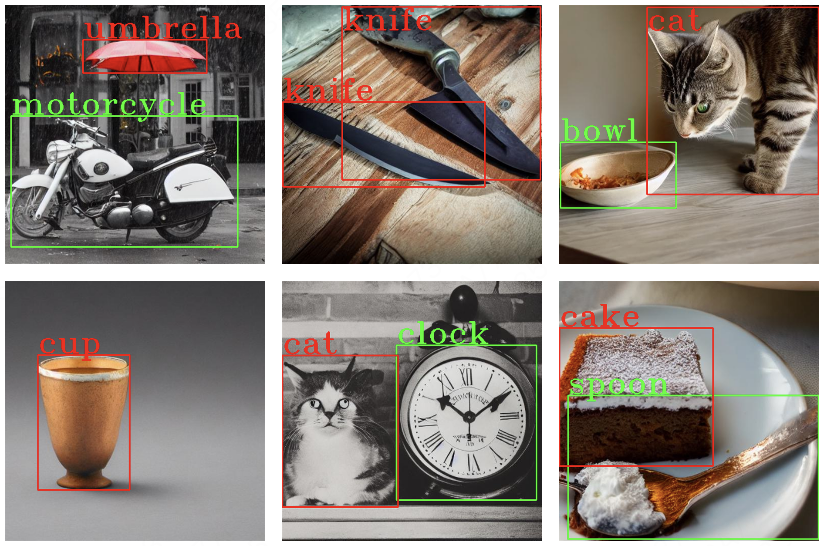}
    \caption{Stable Diffusion + Grounding head w/ Supervised- and \textbf{Self-training}.}
    \label{figure:image-bbox-b}
  \end{subfigure}
  \hfill
  \begin{subfigure}[b]{0.315\textwidth}
    \centering
    \includegraphics[width=\textwidth]{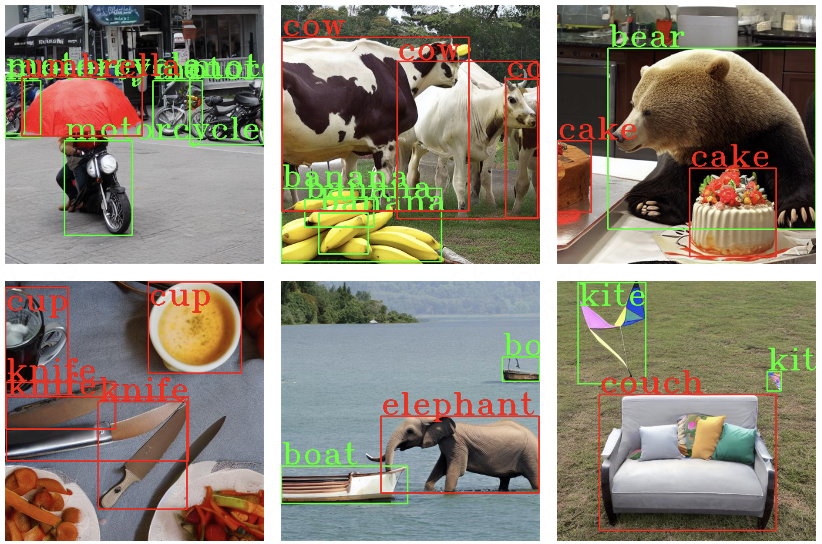}
    \caption{Stable Diffusion w/ \textbf{Fine-tuning} + Grounding head w/ Supervised- and Self-training.}
    \label{figure:image-bbox-c}
  \end{subfigure}
  \vspace{-2mm}
  \caption{Visualization of the synthetic images and bounding-boxes generated from different models. The bounding-boxes with green denote the objects from \textcolor{green}{base} categories, while the ones with red denote the objects from \textcolor{red}{novel} categories.
    }
  \label{figure:image-bbox}
  \vspace{-0.2cm}
\end{figure*}

\vspace{3pt}\noindent \textbf{Training triplets of base categories.}
Following \cite{li2023open}, we apply an automatic pipeline to construct the \{visual feature, bounding-box, text prompt\} triplets, with an object detector trained on base categories from a given dataset~($\mathcal{D}_{\text{real}}$).
In specific, assuming there exists a set of base categories $\{c_{\text{base}}^1, \dots, c_{\text{base}}^N\}$, 
{\em e.g.}, the classes in MS-COCO \cite{lin2014microsoft}.
We first select a random number of base categories to construct a text prompt, {\em e.g.}, `a photograph of [base category1] and [base category2]’, and generate both the visual features and images with our {\em image synthesizer}. 
Then we take an off-the-shelf object detector, 
for example, pre-trained Mask R-CNN~\cite{he2017mask}, to run the inference procedure on the synthetic images, and infer the bounding boxes of the selected categories. To acquire the confident bounding-boxes for training, we use a score threshold $\alpha$ to filter out the bounding-boxes with low confidence (an ablation study on the selection of the score threshold has been conducted in Section~\ref{subsec:ablation}). As a result, an infinite number of training triplets for the given base categories can be constructed by repeating the above operation.

\vspace{3pt}\noindent \textbf{Training loss.} 
We use the constructed training triplets to train the grounding head:
\vspace{-5pt}
\begin{equation}
\label{eq:base-grounding}
\mathcal{L}_{\text{base}} = \sum\limits_{i=1}^N[\mathcal{L}_{\text{cls}}(\hat{s}_{i}, c_{i}) + \mathds{1}_{\{c_i\neq\varnothing\}} \mathcal{L}_{\text{box}}(\hat{b}_i, b_{i})],
\vspace{-3pt}
\end{equation}
where the $i$th prediction ($\hat{s}_{i}$, $\hat{b}_{i}$) from the $N$ object queries is assigned to a ground-truth ($c_{i}$, $b_{i}$) or $\varnothing$ (no object) with bipartite matching.
$\mathcal{L}_{\text{cls}}$ and $\mathcal{L}_{\text{box}}$ denote the classification loss (\eg Focal loss) and box regression loss (\eg L1 loss and GIoU loss), respectively.

\subsubsection{Instance Grounding on Novel Categories}
\label{subsec:self}
Till here, we have obtained a diffusion model with open-vocabulary grounding, which has been only trained with base categories. In this section, we propose to further leverage the synthetic training triplets from a wider range of categories to enhance the alignment for novel/unseen categories. 
Specifically, as shown in Figure~\ref{figure:finetune-b}, we describe a framework that generates the training triplets for novel categories using the grounded diffusion model, and then self-train the grounding head.

\vspace{3pt}
\noindent \textbf{Training triplets of novel categories.} 
We design the text prompts of novel categories, 
{\em e.g.}, ‘a photograph of [novel category1] and [novel category2]’, and pass them through our proposed {\em image synthesizer}, to generate the visual features. To acquire the corresponding bounding-boxes for novel categories, we propose a self-training scheme that takes the above grounding head as the student, and apply a mean teacher~(an exponential moving average (EMA) of the student model) to create pseudo labels for update. In contrast to the widely adopted self-training scheme that takes the image as input, the student and teacher in our case only take the visual features as input, thus {\em cannot} apply data augmentation as for images. Instead, we insert dropout module within each feature enhancer layer and decoder layer in the student.
During training, we run inference~(without dropout module) with teacher model on the visual features to produce bounding boxes, and then use a score threshold $\beta$ to filter out those with low confidence, and use the remaining training triplets $(\mathcal{F}_{i}, \hat{b}_{i}, y_{i}^{\text{novel}})$ to train the student, {\em i.e.}, grounding head.

\vspace{3pt}
\noindent \textbf{Training loss.} 
Now, we can also train the grounding head on the mined triplets of novel categories (that are unseen in the existing real dataset) with the training loss $\mathcal{L}_{\text{novel}}$ defined similar to Eq.~\ref{eq:base-grounding}.
Thus, the total training loss for training the grounding head can be:
$\mathcal{L}_{\text{grounding}} = \mathcal{L}_{\text{base}} + \mathcal{L}_{\text{novel}}$.

\begin{table*} 
\centering
\small
\begin{tabular}{lcccccc}
\toprule
Method & Supervision & Detector & Backbone & AP50$_{\text{all}}^{\text{box}}$ & AP50$_{\text{base}}^{\text{box}}$ & AP50$_{\text{novel}}^{\text{box}}$ \\
\cmidrule(r){1-1}
\cmidrule(r){2-2}
\cmidrule(r){3-4}
\cmidrule(r){5-7}
Detic \cite{zhou2022detecting} & CLIP & Faster R-CNN & R50 & \textcolor{gray!80}{45.0} & \textcolor{gray!80}{47.1} & 27.8 \\
PromptDet \cite{feng2022promptdet} & CLIP & Faster R-CNN & R50 & \textcolor{gray!80}{-} & \textcolor{gray!80}{50.6} & 26.6 \\
BARON \cite{wu2023aligning} & CLIP & Faster R-CNN & R50 & \textcolor{gray!80}{53.5} & \textcolor{gray!80}{60.4} & 34.0 \\
OADP \cite{wang2023object} & CLIP & Faster R-CNN & R50 & \textcolor{gray!80}{47.2} & \textcolor{gray!80}{53.3} & 30.0 \\
ViLD \cite{gu2021open} & CLIP & Mask R-CNN & R50 & \textcolor{gray!80}{51.3} & \textcolor{gray!80}{59.5} & 27.6 \\
F-VLM \cite{kuo2022f} & CLIP & Mask R-CNN & R50 & \textcolor{gray!80}{39.6} & \textcolor{gray!80}{-} & 28.0 \\
RO-ViT \cite{kim2023region} & CLIP & Mask R-CNN & ViT-B \cite{dosovitskiy2020image} & \textcolor{gray!80}{41.5} & \textcolor{gray!80}{-} & 30.2 \\
VLDet \cite{lin2022learning} & CLIP & CenterNet2 \cite{zhou2021probabilistic} & R50 & \textcolor{gray!80}{45.8} & \textcolor{gray!80}{50.6} & 32.0 \\
CxORA \cite{wu2023cora} & CLIP & DAB-DETR \cite{liu2022dab} & R50 & \textcolor{gray!80}{35.4} & \textcolor{gray!80}{35.5} & 35.1 \\
DK-DETR \cite{li2023distilling} & CLIP & Deformable
DETR \cite{zhu2020deformable} & R50 & \textcolor{gray!80}{-} & \textcolor{gray!80}{61.1} & 32.3  \\
EdaDet \cite{shi2023edadet} & CLIP & Deformable DETR \cite{zhu2020deformable} & R50 & \textcolor{gray!80}{52.5} & \textcolor{gray!80}{57.7} & 37.8 \\
\cmidrule(r){1-1}
\cmidrule(r){2-2}
\cmidrule(r){3-4}
\cmidrule(r){5-7}
\name & Stable Diffusion & Faster R-CNN & R50 & \textcolor{gray!80}{52.3} & \textcolor{gray!80}{55.8} & \textbf{42.3} \\ 
\bottomrule
\end{tabular}
\vspace{-2mm}
\caption{Results on open-vocabulary COCO benchmark. 
AP50$_{\text{novel}}^{\text{box}}$ is the main metric for evaluation. Our detector, trained on synthetic dataset from \textbf{InstaGen}, significantly outperforms state-of-the-art CLIP-based approaches on novel categories.
}
\label{table:open-coco}
\vspace{-4mm}
\end{table*}

\vspace{1mm}
\subsection{Training Detector with Synthetic Dataset}
\label{subsec:detector}
In this section, we augment the real dataset~($\mathcal{D}_{\text{real}}$),
with synthetic dataset~($\mathcal{D}_{\text{syn}}$), 
and train popular object detectors, 
for example, Faster R-CNN \cite{ren2015faster} with the standard training loss:
\vspace{-5pt}
\begin{equation}
    \mathcal{L}_{\text{det}} = \mathcal{L}_{\text{rpn\_cls}} + \mathcal{L}_{\text{rpn\_box}} + \mathcal{L}_{\text{det\_cls}} + \mathcal{L}_{\text{det\_box}},
\end{equation}
where $\mathcal{L}_{\text{rpn\_cls}}$, $\mathcal{L}_{\text{rpn\_box}}$ are the classification and box regression losses of region proposal network, and $\mathcal{L}_{\text{det\_cls}}$, $\mathcal{L}_{\text{det\_box}}$ are the classification and box regression losses of the detection head. 
Generally speaking, the synthetic dataset enables to improve the detector's ability from two aspects:
(i) expanding the original data with more categories, 
(ii) improve the detection performance by increasing data diversity.

\vspace{3pt}
\noindent \textbf{Expanding detection categories.}
The grounding head is designed to be open-vocabulary, 
that enables to generate object bounding boxes for novel categories, even though it is trained with a specific set of base categories. This feature enables \textbf{\name}~to construct a detection dataset for any category. Figure~\ref{figure:image-bbox} demonstrates several synthetic images and object bounding boxes for novel categories, {\em i.e.}, the object with red bounding box. We evaluate the effectiveness of training on synthetic dataset through experiments on open-vocabulary detection benchmark.
For more details, please refer to Figure~\ref{figure:first-page}b and Section~\ref{subsec:open-vocabulary}.

\vspace{3pt}
\noindent \textbf{Increasing data diversity.}
The base diffusion model is trained on a large corpus of image-caption pairs, that enables to generate diverse images. 
Taking advantage of such capabilities, \textbf{\name}~is capable of generating dataset with diverse images and box annotations,
which can expand the original dataset, {\em i.e.}, increase the data diversity and  improve detection performance, particularly in data-sparse scenarios. We conducted experiments with varying proportions of COCO~\cite{lin2014microsoft} images as available real data, and show the effectiveness of  training on synthetic dataset when the number of real-world images is limited. 
We refer the readers for more details in Section \ref{subsec:data-sparse}, and results in Figure~\ref{figure:first-page}c.

\begin{table*}
\small
\begin{minipage}[t]{.32\linewidth}
\centering
\setlength\tabcolsep{3.3pt}
\begin{tabular}[t]{cccccc}
\toprule
InstaGen & 10\% & 25\% & 50\% & 75\% & 100\% \\
\cmidrule(r){1-1}
\cmidrule(r){2-6}
\xmark & 23.3 & 29.5 & 34.1 & 36.1 & 37.5 \\
\cmark & 28.5 & 32.6 & 35.8 & 37.3 & 38.5 \\
\bottomrule
\end{tabular}
\vspace{-1mm}
\captionof{table}{Results on data-sparse object detection. 
We employ Faster R-CNN with the ResNet-50 backbone as the default object detector and evaluate its performance using the AP metric on MS COCO benchmark. 
Please refer to the text for more details.
}
\vspace{-1mm}
\label{table:scarce}
\end{minipage} 
\hfill
\begin{minipage}[t]{.65\linewidth}
\centering
\setlength\tabcolsep{4.5pt}
\begin{tabular}[t]{lccccc}
\toprule
Method & Supervision & Detector & Extra Data & Object365 & LVIS \\
\cmidrule(r){1-1}
\cmidrule(r){2-2}
\cmidrule(r){3-3}
\cmidrule(r){4-4}
\cmidrule(r){5-6}
Gao \etal \cite{gao2022open} & CLIP & CenterNet2 & \cmark & 6.9 & 8.0 \\
VL-PLM \cite{zhao2022exploiting} & CLIP & Mask R-CNN & \cmark & 10.9 & 22.2 \\
\cmidrule(r){1-1}
\cmidrule(r){2-2}
\cmidrule(r){3-3}
\cmidrule(r){4-4}
\cmidrule(r){5-6}
\name & Stable Diffusion & Faster R-CNN & \xmark & \textbf{11.4} & \textbf{23.3} \\ 
\bottomrule
\end{tabular}
\vspace{-2mm}
\captionof{table}{Results on generalizing COCO-base to Object365 and LVIS. All detectors utilize the ResNet-50 backbone. The evaluation protocol follows \cite{gao2022open} and reports AP50. 
Extra data refers to an additional dataset that encompasses objects from the categories within the target dataset. 
In both experiments, the extra data consists of all the images from COCO, which has covered the majority of categories in Object365 and LVIS.
} 
\vspace{-1mm}
\label{table:unseen_dataset}
\end{minipage}
\vspace{-0.3cm}
\end{table*}

\begin{table} 
\small
\centering
\begin{tabular}{cccccc}
\toprule
G-head & ST & FT & AP50$_{\text{all}}^{\text{box}}$ & AP50$_{\text{base}}^{\text{box}}$ & AP50$_{\text{novel}}^{\text{box}}$ \\
\cmidrule(r){1-3}
\cmidrule(r){4-6}
\cmark &  &  & 50.6 & 55.3 & 37.1 \\
\cmark & \cmark &  & 51.1 & 55.0 & 40.3 \\
\cmark & \cmark  & \cmark & 52.3 & 55.8 & 42.3 \\
\bottomrule
\end{tabular}
\vspace{-2mm}
\caption{The effectiveness of the proposed components. G-head, ST and FT refer to the grounding head, self-training the grounding head and fine-tuning SDM, respectively.}
\label{table:componets}
\vspace{-5mm}
\end{table}

\vspace{-1mm}
\section{Experiment}
\label{sec:exp}

In this section, we use the proposed \textbf{\name}~to construct synthetic dataset for training object detectors, {\em i.e.}, generating images with the corresponding bounding boxes. Specifically, we present the implementation details in Section~\ref{subsec:implementation}. 
To evaluate the effectiveness of the synthetic dataset for training object detector, we consider three protocols: open-vocabulary object detection~(Section~\ref{subsec:open-vocabulary}), 
data-sparse object detection~(Section~\ref{subsec:data-sparse}) 
and cross-dataset object detection~(Section~\ref{subsec:cross-data}). 
Lastly, we conduct ablation studies on the effectiveness of the proposed components and the selection of hyper-parameters~(Section~\ref{subsec:ablation}).

\subsection{Implementation details}
\label{subsec:implementation}
\noindent \textbf{Network architecture.}
We build {\em image synthesizer} from the pre-trained Stable Diffusion v1.4~\cite{rombach2022high}, and use the CLIP text encoder~\cite{radford2021learning} to get text embedding for the category name. The channel compression layer maps the dimension of visual features to 256, which is implemented with a 3$\times$3 convolution. For simplicity, the feature enhancer, language-guided query selection module and cross-modality decoder are designed to the same structure as the ones in~\cite{liu2023grounding}. 
The number of the object queries is set to 900.

\vspace{3pt}\noindent \textbf{Constructing image synthesizer.} 
In our experiments, we first fine-tune the stable diffusion model on a real detection dataset, {\em e.g.}, the images of base categories. 
During training, the text encoder of CLIP is kept frozen, while the remaining components are trained for 6 epochs with a batch size of 16 and a learning rate of 1e-4.

\vspace{3pt}\noindent \textbf{Instance grounding module.} 
We start by constructing the training triplets using base categories {\em i.e.}, the categories present in the existing dataset. 
The text prompt for each triplet is constructed by randomly selecting one or two categories. The regional visual features are taken from the {\em image synthesizer} time step $t=1$, and the oracle ground-truth bounding boxes are obtained using a Mask R-CNN model trained on base categories, as explained in Section~\ref{subsec:head}.

Subsequently, we train the instance grounding module with these training triplets for 6 epochs, with a batch size of 32. 
In the 6th epoch, we transfer the weights from the student model to the teacher model, and proceed to train the student for an additional 6 epochs. During this training, the student receives supervised training on the base categories and engages in self-training on novel categories, and the teacher model is updated using exponential moving average (EMA) with a momentum of 0.999. The initial learning rate is set to 1e-4 and is subsequently reduced by a factor of 10 at the 11th epoch, and the score thresholds $\alpha$ and $\beta$ are set to 0.8 and 0.4, respectively.

\vspace{3pt}\noindent \textbf{Training object detector on combined dataset.} In our experiment, we train an object detector~(Faster R-CNN~\cite{ren2015faster}) with ResNet-50~\cite{he2016deep} as backbone, on a combination of the existing real dataset and the synthetic dataset. 
Specifically, for synthetic dataset, we randomly select one or two categories at each iteration, construct the text prompts, and feed them as input to generates images along with the corresponding bounding boxes with $\beta$ of 0.4. Following the standard implementation~\cite{ren2015faster}, the detector is trained for 12 epochs (1$\times$ learning schedule) unless specified. The initial learning rate is set to 0.01 and then reduced by a factor of 10 at the 8th and the 11th epochs.

\begin{table*}[!htb]
\small
\begin{minipage}[t]{.32\linewidth}
\setlength\tabcolsep{3pt}
\centering
\begin{tabular}{lccc}
\toprule
\#Images & AP50$_{\text{all}}^{\text{box}}$ & AP50$_{\text{base}}^{\text{box}}$ & AP50$_{\text{novel}}^{\text{box}}$ \\
\cmidrule(r){1-1}
\cmidrule(r){2-4}
1000 & 51.6 & 55.9 & 39.7 \\
2000 & 51.7 & 55.4 & 41.1 \\
3000 & 52.3 & 55.8 & 42.3 \\
\bottomrule
\end{tabular}
\vspace{-0.2cm}
\captionof{table}{Number of generated images.}
\label{table:image-number}
\end{minipage} 
\hfill
\hspace{13pt}
\begin{minipage}[t]{.3\linewidth}
\centering
\setlength\tabcolsep{3.5pt}
\begin{tabular}{lccc}
\toprule
$\alpha$ & AP50$_{\text{all}}^{\text{box}}$ & AP50$_{\text{base}}^{\text{box}}$ & AP50$_{\text{novel}}^{\text{box}}$ \\
\cmidrule(r){1-1}
\cmidrule(r){2-4}
0.7 & 51.3 & 55.1 & 40.6 \\
0.8 & 52.3 & 55.8 & 42.3 \\
0.9 & 51.8 & 55.6 & 41.1 \\
\bottomrule
\end{tabular}
\vspace{-0.2cm}
\captionof{table}{$\alpha$ for bounding-box filtration.}
\label{table:alpha}
\end{minipage}
\hfill
\hspace{13pt}
\begin{minipage}[t]{.3\linewidth}
\centering
\setlength\tabcolsep{3.5pt}
\begin{tabular}{lccc}
\toprule
$\beta$ & AP50$_{\text{all}}^{\text{box}}$ & AP50$_{\text{base}}^{\text{box}}$ & AP50$_{\text{novel}}^{\text{box}}$ \\
\cmidrule(r){1-1}
\cmidrule(r){2-4}
0.3 & 46.4 & 53.3 & 26.9 \\
0.4 & 52.3 & 55.8 & 42.3 \\
0.5 & 51.2 & 55.4 & 39.2 \\
\bottomrule
\end{tabular}
\vspace{-0.2cm}
\captionof{table}{$\beta$ for bounding-box filtration.}
\label{table:beta}
\end{minipage}
\vspace{-0.5cm}
\end{table*}

\subsection{Open-vocabulary object detection} 
\label{subsec:open-vocabulary}
\noindent \textbf{Experimental setup.} 
Following the previous works~\cite{feng2022promptdet,zhao2022exploiting}, 
we conduct experiments on the open-vocabulary COCO benchmark, 
where 48 classes are treated as base categories, and 17 classes as the novel categories. 
More results for LVIS can be found in the \textbf{supplementary material}.
To train the grounding head, we employ 1250 synthetic images per category per training epoch. While for training the object detector, we use 3000 synthetic images per category,
along with the original real dataset for base categories.
The object detector is trained with input size of $800\times800$ and scale jitter. The performance is measured by COCO Average Precision at an Intersection over Union of 0.5 (AP50).

\vspace{3pt}\noindent \textbf{Comparison to SOTA.}
As shown in Table~\ref{table:open-coco}, 
we evaluate the performance by comparing with existing CLIP-based open-vocabulary object detectors. It is clear that our detector trained on synthetic dataset from \textbf{\name}~outperforms existing state-of-the-art approaches significantly, {\em i.e.}, around $+$5AP improvement over the second best. In essence, through the utilization of our proposed open-vocabulary grounding head, \textbf{\name}~is able to generate detection data for novel categories, 
enabling the detector to attain exceptional performance. 
To the best of our knowledge, this is the first work that applies generative diffusion model for dataset synthesis, to tackle open-vocabulary object detection, and showcase its superiority in this task.

\subsection{Data-sparse object detection}
\label{subsec:data-sparse}
\noindent \textbf{Experimental setup.} 
Here, we evaluate the effectiveness of synthetic dataset in data-spare scenario,
by varying the amount of real data. We randomly select subsets comprising 10\%, 25\%, 50\%, 75\% and 100\% of the COCO training set,
this covers all COCO categories.
These subsets are used to fine-tune stable diffusion model for constructing {\em image synthesizer}, and train a Mask R-CNN for generating oracle ground-truth bounding boxes in synthetic images. We employ 1250 synthetic images per category to train a Faster R-CNN in conjunction with the corresponding COCO subset. The performance is measured by Average Precision~\cite{lin2014microsoft}.

\vspace{3pt}\noindent \textbf{Comparison to baseline.}
As shown in Table~\ref{table:scarce}, the Faster R-CNN trained with synthetic images achieves consistent improvement across various real training data budgets. Notably, as the availability of real data becomes sparse,
synthetic dataset plays even more important role for performance improvement, for instance, it improves the detector by +5.2 AP (23.3$\rightarrow$28.5 AP) when only 10\% real COCO training subset is available.

\subsection{Cross-dataset object detection}
\label{subsec:cross-data}
\noindent \textbf{Experimental setup.} 
In this section, we assess the effectiveness of synthetic data on a more challenging task, namely cross-dataset object detection.  
Following~\cite{zhao2022exploiting}, we evaluate the COCO-trained model on two unseen datasets: Object365~\cite{shao2019objects365} and LVIS~\cite{gupta2019lvis}. Specifically, we consider the 48 classes in the open-vocabulary COCO benchmark as the source dataset, while Object365 (with 365 classes) and LVIS (with 1203 classes) serve as the target dataset. 
When training the instance grounding module, we acquire 1250 synthetic images for base categories from the source dataset, and 100 synthetic images for the category from the target dataset at each training iteration. In the case of training the object detector, we employ 500 synthetic images per category from the target dataset for each training iteration. The detector is trained with input size of $1024\times1024$ and scale jitter~\cite{zhao2022exploiting}.

\vspace{3pt}\noindent \textbf{Comparison to SOTA.} 
The results presented in Table \ref{table:unseen_dataset} demonstrate that the proposed \textbf{\name}~achieves superior performance in generalization from COCO-base to Object365 and LVIS, when compared to CLIP-based methods such as~\cite{gao2022open,zhao2022exploiting}. It is worth noting that CLIP-based methods require the generation of pseudo-labels for the categories from the target dataset on COCO images, and subsequently train the detector using these images. These methods necessitate a dataset that includes objects belonging to the categories of the target dataset. In contrast, \textbf{\name}~possesses the ability to generate images featuring objects of any category without the need for additional datasets, thereby enhancing its versatility across various scenarios.

\subsection{Ablation study}
\label{subsec:ablation}
To understand the effectiveness of the proposed components,
we perform thorough ablation studies on the open-vocabulary COCO benchmark~\cite{lin2014microsoft}, investigating the effect of fine-tuning stable diffusion model, training instance grounding module, self-training on novel categories. Additionally, we investigate other hyper-parameters by comparing the effectiveness of synthetic images and different score thresholds for base and novel categories.

\vspace{3pt}\noindent \textbf{Fine-tuning diffusion model.}
We assess the effectiveness of fine-tuning stable diffusion model, 
and its impact for synthesizing images for training object detector. Figure~\ref{figure:image-bbox-c} illustrates that \textbf{\name}~is capable of generating images with more intricate contexts, featuring multiple objects, small objects, and occluded objects. Subsequently, we employed these generated images to train Faster R-CNN for object detection. The results are presented in Table~\ref{table:componets}, showing that {\em image synthesizer} from fine-tuning stable diffusion model delivers improvement detection performance by 2.0 AP (from 40.3 to 42.3 AP).

\vspace{3pt}\noindent \textbf{Instance grounding module.}
To demonstrate the effectiveness of the grounding head in open-vocabulary scenario, we exclusively train it on base categories. Visualization examples of the generated images are presented in Figure~\ref{figure:image-bbox-a}. These examples demonstrate that the trained grounding head is also capable of predicting bounding boxes for instances from novel categories. Leveraging these generated images to train the object detector leads to a 37.1 AP on novel categories, 
surpassing or rivaling all existing state-of-the-art methods, as shown in Table~\ref{table:open-coco} and Table~\ref{table:componets}.

\vspace{3pt}\noindent \textbf{Self-training scheme.}
We evaluate the performance after self-training the grounding head with novel categories. As shown in Table~\ref{table:componets}, training Faster R-CNN with the generated images of novel categories, leads to a noticeable enhancement in detection performance, increasing from 37.1 to 40.3 AP. Qualitatively, it also demonstrates enhanced recall for novel objects after self-training, as shown in Figure~\ref{figure:image-bbox-b}.

\vspace{3pt}\noindent \textbf{Number of synthetic images.}
We investigate the performance variation while increasing the number of the generated images per category for detector training. As shown in Table~\ref{table:image-number}, when increasing the number of generated images from 1000 to 3000, the detector's performance tends to be increasing monotonically, from 39.7 to 42.3 AP on novel categories, showing the scalability of the proposed training mechanism.

\vspace{3pt}\noindent \textbf{Score thresholds for bounding box filtration.}
We compare the performance with different score thresholds $\alpha$ and $\beta$ for filtering bounding boxes on base categories and novel categories, respectively. From the experiment results in Table~\ref{table:alpha}, 
we observe that the performance is not sensitive to the value of $\alpha$, 
and $\alpha=0.8$ yields the best performance. 
The experimental results using different $\beta$ are presented in Table~\ref{table:beta}. With a low score threshold ($\alpha=0.3$), 
there are still numerous inaccurate bounding boxes remaining, 
resulting in an AP of 26.9 for novel categories. by increasing $\beta$ to 0.4, numerous inaccurate bounding boxes are filtered out, resulting in optimal performance. Hence, we set $\alpha=0.8$ and $\beta=0.4$ in our experiments.

\vspace{-1mm}
\section{Limitation}
Using synthetic or artificially generated data in training AI algorithms is a burgeoning practice with significant potential. It can address data scarcity, privacy, and bias issues. 
However, there remains two limitations for training object detectors with synthetic data, 
(i) synthetic datasets commonly focus on clean, 
isolated object instances, which limits the exposure of the detector to the complexities and contextual diversity of real-world scenes, such as occlusions, clutter, varied environmental factors, deformation, therefore, models trained on synthetic data struggle to adapt to real-world conditions, affecting their overall robustness and accuracy,
(ii) existing diffusion-based generative model also suffers from long-tail issue,
that means the generative model struggles to generate images for objects of rare categories, resulting in imbalanced class representation during training and reduced detector performance for less common objects.

\vspace{-1mm}
\section{Conclusion}
\label{sec:conclusion}
This paper proposes a dataset synthesis pipeline, 
termed as \textbf{\name}, that enables to generate images with object bounding boxes for arbitrary categories, acting as a annotation-free approach for constructing large-scale synthetic dataset to train object detector. We have conducted thorough experiments to show the effectiveness of training on synthetic data, on improving detection performance, or expanding the number of detection categories. Significant improvements have been shown in various detection scenarios, including open-vocabulary~($+4.5$~AP) and data-sparse~($+1.2\sim5.2$~AP) detection.

{
    \small
    \bibliographystyle{ieeenat_fullname}
    \bibliography{main}
}

\clearpage
\setcounter{page}{1}
\setcounter{table}{0}
\setcounter{figure}{0}
\setcounter{section}{0}
\renewcommand{\thetable}{S\arabic{table}}
\renewcommand{\thefigure}{S\arabic{figure}}
\renewcommand{\thesection}{S\arabic{section}}

\maketitlesupplementary

In this supplementary document, we present the experimental results of the LVIS-OVD benchmark in Section \ref{sec:lvis-ovd}. Additionally, we perform an ablation study to evaluate the coupling between the diffusion model and the grounding head in Section \ref{sec:coupling}. Furthermore, we evaluate the quality of the pseudo-labels generated by the grounding head in Section \ref{sec:labels}. Lastly, we provide more qualitative results in Section \ref{sec:qualitative}.

\section{Open-vocabulary setting on LVIS}
\label{sec:lvis-ovd}
\noindent \textbf{Experimental setup.} 
We conduct experiments on the LVIS-OVD benchmark. The latest LVIS v1.0~\cite{gupta2019lvis} consists of 1203 categories, each with bounding box and instance mask annotations. The categories are divided into three groups based on the number of images in which each category appears in the training set: rare (1-10 images), common (11-100 images), and frequent (more than 100 images). In line with the problem setting in ViLD~\cite{gu2021open} and Detic~\cite{zhou2022detecting}, we treat the frequent and common classes as base categories, while considering the rare classes as novel categories. For evaluation on LVIS v1.0 $minival$ set,
we mainly consider the mask Average Precision for novel categories, 
{\em i.e.}~AP$_{\text{novel}}$. However, to complete the AP metric, 
we also report AP$_{\text{c}}$~(for common classes), AP$_{\text{f}}$~(for frequent classes) and AP (for all classes). 

Similar to PromptDet \cite{feng2022promptdet}, we enhance the prompt template by incorporating a more detailed description to mitigate lexical ambiguity, particularly for the rare classes in LVIS. It should be noted that the description can be easily extracted from the metadata of the dataset. Consequently, the text prompt for the selected categories is generated as follows: `a photograph of [category1 name] ([category1 description]) and [category2 name] ([category2 description])'.
During the training of the grounding head, we utilize 500 synthetic images per category per training epoch. In addition, for the training of the object detector, we employ 250 synthetic images per category per training epoch and conduct 24 epochs of training.

\vspace{3pt}\noindent \textbf{Comparison to SOTA.} 
We conduct a comparison with the existing CLIP-based open-vocabulary object detectors using the Mask-RCNN model with ResNet-50, as shown in Table \ref{ex:lvis-ovd}. The results indicate that our detector, trained on synthetic dataset from \textbf{\name}, achieves comparable or improved performance over existing CLIP-based methods.

\section{Tight coupling \vs Loose coupling}
\label{sec:coupling}
To generate high-quality bounding-boxes for the synthetic images, we have designed a tight coupling between the diffusion model and the instance-level grounding head, namely, the grounding head predicts the bounding-boxes based on the SDM’s internal representation. To demonstrate the effectiveness of the tight coupling design, we compare it with a loose coupling design. For the latter, we train an open-vocabulary detector (\ie ResNet-101 + instance level grounding head) on the synthetic images with base categories, and generate pseudo-labels for novel categories. When training detectors on such synthetic dataset, it gives 31.9 AP on novel categories on the COCO-OVD benchmark, 10.4 AP lower than tight coupling, showing the benefits of rich semantic and positional information encoded in SDM’s visual features.

\section{Quality of Pseudo-labels}
\label{sec:labels}
Here we evaluate the quality of the pseudo-labels generated by the proposed grounding head.
We adopt two metrics to assess their quality: (i) Detector AP and (ii) Precision and Recall. For Detector AP, we leverage the pre-trained Mask-RCNN model on the COCO dataset to generate ground truths (GTs) for the synthetic images, and then compute the AP of the pseudo labels derived from the teacher model. In the case of Precision and Recall, we randomly select and annotate 200 synthetic images, then calculate the precision and recall of their pseudo-labels. As shown in Table \ref{table:pseudo-gts}, after self-training on novel categories, the quality of the pseudo-labels can be significantly improved in terms of Detector AP (70.2\%$\rightarrow$79.7\%), Precision (87.9\%$\rightarrow$89.1\%) and Recall (68.3\%$\rightarrow$90.0\%).

\begin{table}[t] 
\centering
\begin{tabular}{ccccc}
\toprule

$\mathcal{L}_{\text{base}}$ & $\mathcal{L}_{\text{novel}}$ & Detector AP & Precision & Recall \\
\cmidrule(r){1-2}
\cmidrule(r){3-3}
\cmidrule(r){4-5}
\cmark & & 70.2 & 87.9 & 68.3\\
\cmark & \cmark & 79.7 & 89.1 & 90.0 \\
\bottomrule
\end{tabular}
\caption{The quality of the pseudo-labels.}
\label{table:pseudo-gts}
\end{table}

\section{Qualitative Results}
\label{sec:qualitative}
\begin{figure*}
		\centering
		\includegraphics[width=\textwidth]{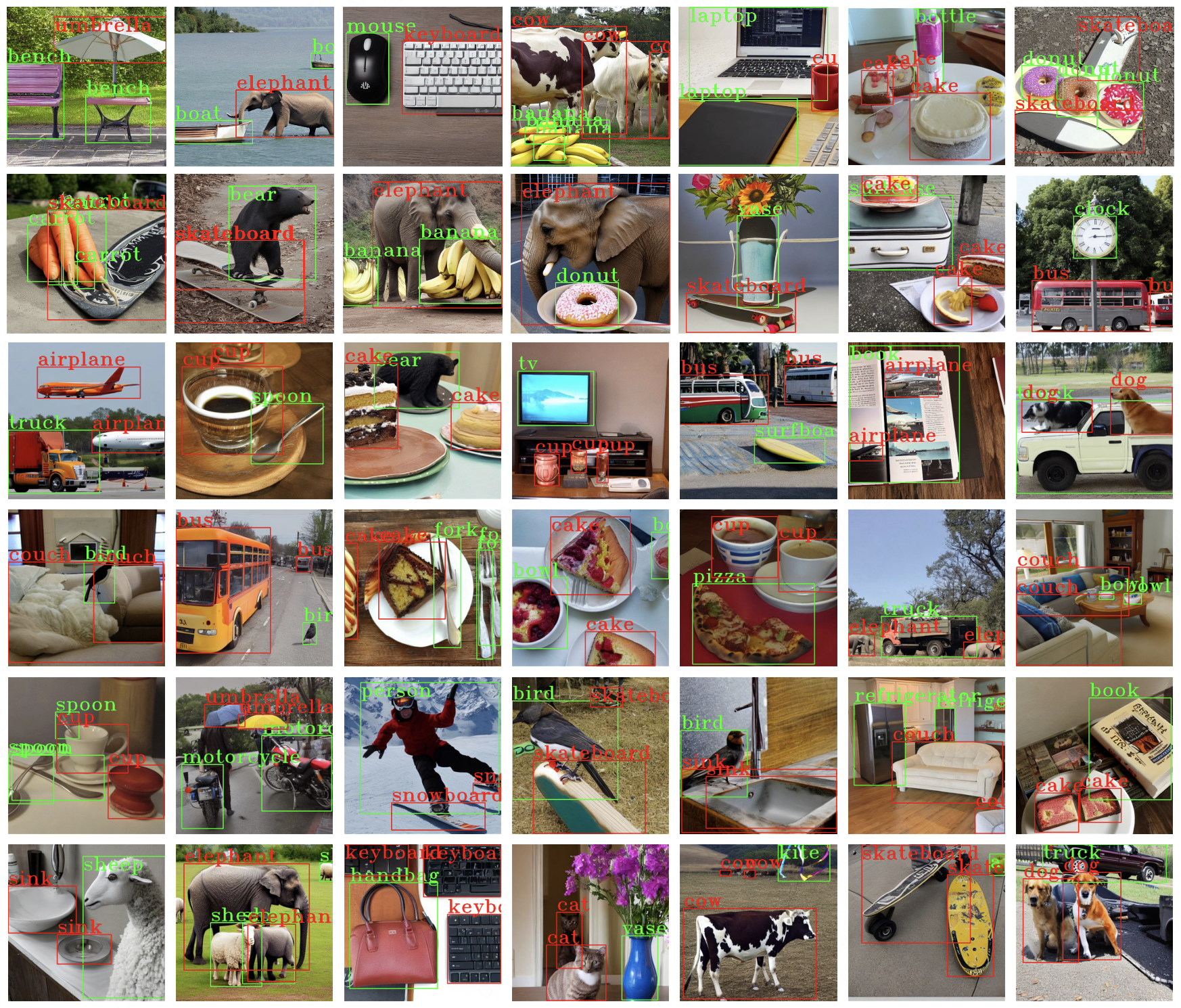}
		\caption{Qualitative results generated by our InstaGen. The bounding-boxes with green denote the objects from \textcolor{green}{base} categories, while the ones with red denote the objects from \textcolor{red}{novel} categories.}
		\label{figure:demo_gen}
\end{figure*}

\begin{figure*}
		\centering
		\includegraphics[width=\textwidth]{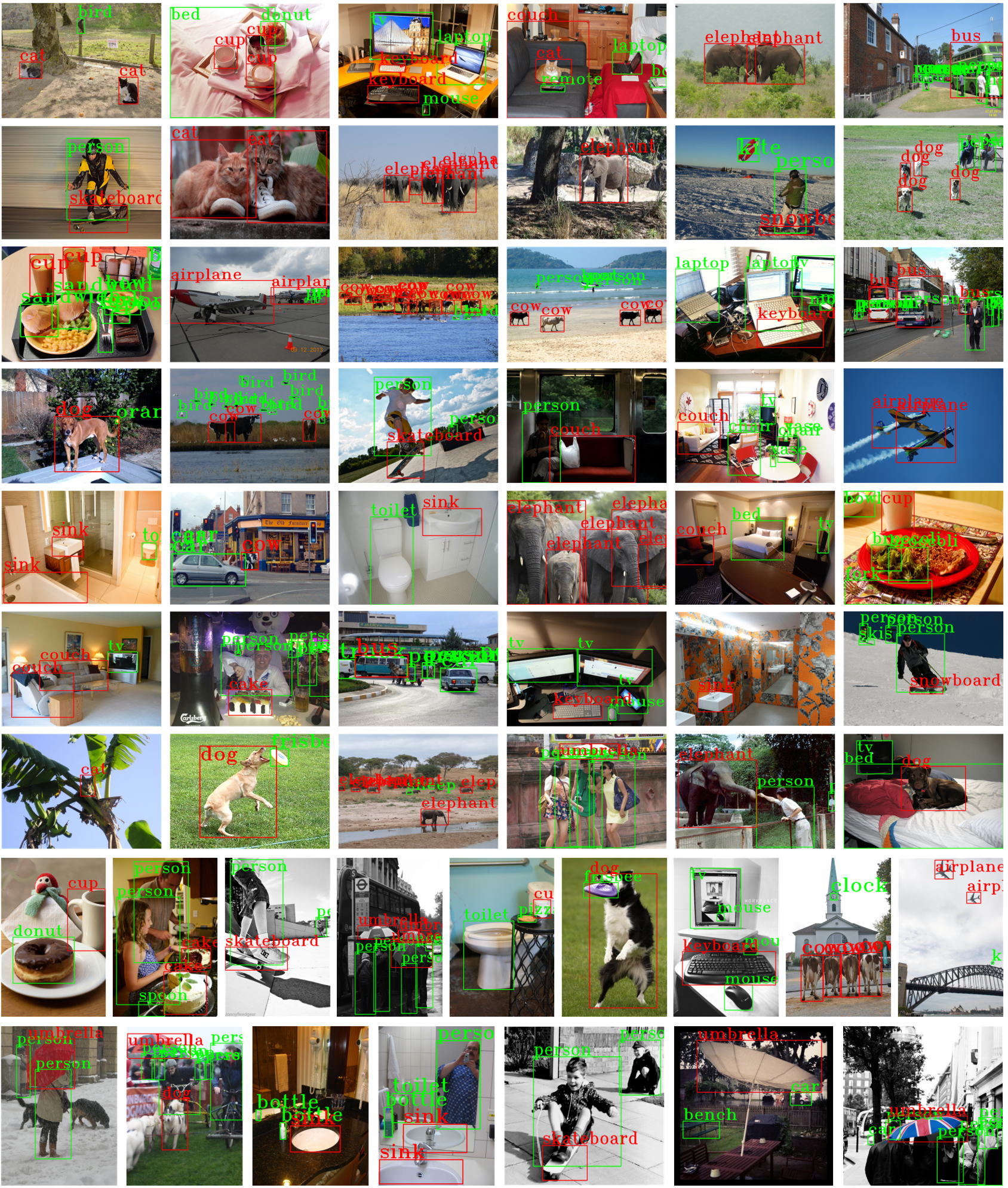}
		\caption{Qualitative results from our Faster R-CNN trained with the synthetic images from InstaGen on COCO validation set. The bounding-boxes with green denote the objects from \textcolor{green}{base} categories, while the ones with red denote the objects from \textcolor{red}{novel} categories.}
		\label{figure:demo_det}
\end{figure*}

We show more qualitative results generated by our InstaGen in Figure \ref{figure:demo_gen}. Without any manual annotations, InstaGen can generate high-quality images with object bounding-boxes of novel categories. In Figure \ref{figure:demo_det}, we further show the qualitative results predicted by the Faster R-CNN trained with the synthetic images form InstaGen on COCO validation set. The detector can now accurately localize and recognize the objects from novel categories.

\begin{table*}[h]
\vspace{-3mm}
\begin{center}
\setlength\tabcolsep{5pt}
\begin{tabular}{lcccccccc}
\toprule
Method & Supervision & Detector & Backbone & Input Size & AP & AP$_{\text{c}}$ & AP$_{\text{f}}$ & AP$_{\text{novel}}$ \\
\cmidrule(r){1-1}
\cmidrule(r){2-2}
\cmidrule(r){3-5}
\cmidrule(r){6-9}
ViLD-ens. \cite{gu2021open} & CLIP & Mask R-CNN & R50 & 1024$\times$1024 & \textcolor{gray!80}{25.5} & \textcolor{gray!80}{24.6} & \textcolor{gray!80}{30.3} & 16.6  \\
Detic \cite{zhou2022detecting} & CLIP & Mask R-CNN & R50 & 1024$\times$1024 & \textcolor{gray!80}{26.8} & \textcolor{gray!80}{26.3} & \textcolor{gray!80}{31.6} & 17.8 \\
F-VLM \cite{kuo2022f} & CLIP & Mask R-CNN & R50 & 1024$\times$1024 & \textcolor{gray!80}{24.2} & \textcolor{gray!80}{-} & \textcolor{gray!80}{-} & 18.6 \\
PromptDet \cite{feng2022promptdet} & CLIP & Mask R-CNN & R50 & 800$\times$800 & \textcolor{gray!80}{21.4} & \textcolor{gray!80}{18.5} & \textcolor{gray!80}{25.8} & 19.0 \\ 
DetPro \cite{du2022learning} & CLIP & Mask R-CNN & R50 & 800$\times$800 & \textcolor{gray!80}{25.9} & \textcolor{gray!80}{25.6} & \textcolor{gray!80}{28.9} & 19.8 \\
BARON \cite{wu2023aligning} & CLIP & Mask R-CNN & R50 & 800$\times$800 & \textcolor{gray!80}{25.1} & \textcolor{gray!80}{24.4} & \textcolor{gray!80}{28.9} & 18.0 \\
BARON \cite{wu2023aligning}$^{\dag}$ & CLIP & Mask R-CNN & R50 & 800$\times$800 & \textcolor{gray!80}{27.6} & \textcolor{gray!80}{27.6} & \textcolor{gray!80}{29.8} & 22.6 \\
\cmidrule(r){1-1}
\cmidrule(r){2-2}
\cmidrule(r){3-5}
\cmidrule(r){6-9}
InstaGen & Stable Diffusion & Mask R-CNN & R50 & 800$\times$800 & \textcolor{gray!80}{23.0} & \textcolor{gray!80}{20.6} & \textcolor{gray!80}{27.1} & 20.3 \\
\bottomrule
\end{tabular}
\vspace{-4mm}
\end{center}
\caption{Results on open-vocabulary LVIS benchmark. $^{\dag}$ indicates using ensembling strategy for classification scores and learned prompts for the category’s names.}
\label{ex:lvis-ovd}
\end{table*}

\end{document}